\documentclass{article}

\usepackage[preprint]{neurips_2024}

\usepackage[utf8]{inputenc} 
\usepackage[T1]{fontenc}    
\usepackage{hyperref}       
\usepackage{url}            
\usepackage{booktabs}       
\usepackage{amsfonts}       
\usepackage{nicefrac}       
\usepackage{microtype}      
\usepackage{xcolor}         
\usepackage{graphicx} 
\usepackage{amsmath}
\usepackage{amssymb}
\usepackage{mathtools}
\usepackage{amsthm}
\usepackage{subcaption}
\usepackage{multirow}

\title{Efficient and Generalized end-to-end Autonomous Driving System with Latent Deep Reinforcement Learning and Demonstrations}

\author{%
  Zuojin Tang$^{1,2}$ \\
  \And
  Xiaoyu Chen$^{3}$ \\
  \And
  Yongqiang Li$^{4}$ \\
  \And
  Jianyu Chen$^{1,3*}$ \\
  \And
  \\
  $^1$Shanghai Qizhi Institute \\
  $^2$College of Computer Science and Technology, Zhejiang University \\
  $^3$Institute for Interdisciplinary Information Sciences, Tsinghua University \\
  $^4$Mogo Auto Intelligence and Telematics Information Technology Co., Ltd \\
  $^*$Corresponding to: jianyuchen@tsinghua.edu.cn
}

\begin{document}

\maketitle
\begin{abstract}
An intelligent driving system should dynamically formulate appropriate driving strategies based on the current environment and vehicle status while ensuring system security and reliability. However, methods based on reinforcement learning and imitation learning often suffer from high sample complexity, poor generalization, and low safety. To address these challenges, this paper introduces an efficient and generalized end-to-end autonomous driving system (EGADS) for complex and varied scenarios. The RL agent in our EGADS combines variational inference with normalizing flows, which are independent of distribution assumptions. This combination allows the agent to capture historical information relevant to driving in latent space effectively, thereby significantly reducing sample complexity. Additionally, we enhance safety by formulating robust safety constraints and improve generalization and performance by integrating RL with expert demonstrations. Experimental results demonstrate that, compared to existing methods, EGADS significantly reduces sample complexity, greatly improves safety performance, and exhibits strong generalization capabilities in complex urban scenarios. Particularly, we contributed an expert dataset collected through human expert steering wheel control, specifically using the G29 steering wheel. Our code is available: \textit{https://github.com/Mark-zjtang/EGADS?tab=readme-ov-file}.
\end{abstract}
\begin{figure*}[!htp]
	\centering{\includegraphics[width=12cm]{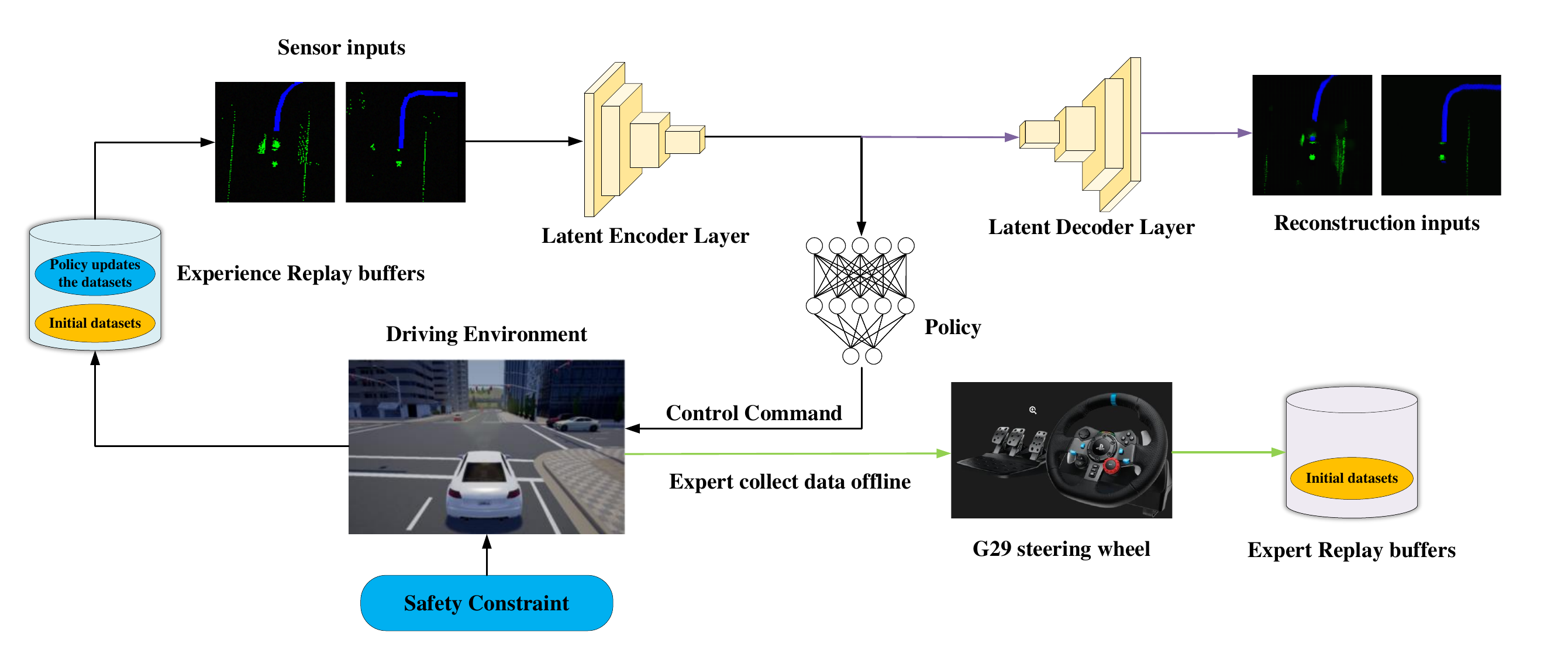}}
	\caption{Overview of the efficient and generalized end-to-end autonomous driving system with
latent deep reinforcement learning and demonstrations. } \label{image1}
\end{figure*}

\section{Introduction}

An intelligent autonomous driving systems must be able to handle complex road geometry and topology, complex multi-agent interactions with dense surrounding dynamic objects, and accurately follow the planning and obstacle avoidance. 
Current, autonomous driving systems in industry are mainly using a highly modularized hand-engineered approach, for example, perception, localization, behavior prediction, decision making and motion control, etc. \cite{thrun2006stanley} and \cite{urmson2008autonomous}. Particularly, the autonomous driving decision making systems are focusing on the non-learning model-based methods, which often requires to manually design a driving policy \cite{gonzalez2015review} and \cite{paden2016survey}. However, the manually designed policy could have two several weaknesses: 1) Accuracy. The driving policy of human heuristics and pre-training model can be suboptimal, which will lead to either conservative or aggressive driving policies. 2) Generality. For different scenarios and complicated tasks, we might need to be redesigned the model policy manually for each new scenario. 

To solve those problems, existing works such as imitation learning (IL) is most popular approach, which can learn a driving policy by collecting the expert driving data. However, those methods can suffer from the following shortcomings for imitation learning: (1) High training cost and sample complexity. (2) Conservation. Due to the collect driving data from the human expert, which can only learn driving skills that are demonstrated in the datasets. (3) Limitation of driving performance. What’s more, the driving policy based on reinforcement learning (RL) is also popular method in recent years, which can automatically learn and explore without any human expert data in various kinds of different driving cases, and it is possible to have a better performance than imitation learning.  
However, the existing methods also have some weakness:
(1) Existing methods in latent space are based on specific distribution assumptions, whereas distributions in the real world tend to be more flexible, resulting in a failure to learn more precisely about belief values. (2) High costs of learning and exploration. (3) The safety and generalization of intelligent vehicles need further improvement.

Combining the advantages of RL and IL, the demonstration of enhanced RL is not only expected to accelerate the initial learning process, but also gain the potential of experts beyond performance. In this paper, we introduce an efficient and generalized end-to-end autonomous driving system (EGADS) for complex and varied scenarios. The RL agent in our EGADS combines variational inference with normalizing flows independent of distribution assumptions, allowing it to sufficiently and flexibly capture historical information useful for driving in latent space, thereby significantly reducing sample complexity. In addition, unlike traditional methods that constrain policy actions directly, we integrate safety constraints into the reward function, which allows the agent to consider safety during training, thereby improving its robustness and generalization. To further increase the upper limit of the overall system, we further enhance the RL search process with a dataset of human experts. In particular, we contributed a dataset of human experts to driving by driving the G29 steering wheel. The experimental results show that compared with the existing methods, our EGADS greatly improves the safety performance, shows strong generalization ability in multiple test maps, and significantly reduces the sample complexity. In summary, our contributions are:
\begin{itemize}
\item[•]We present an EGADS framework designed for complex and varied scenarios.
\item[•]The RL agent in EGADS uses variational inference with normalizing flows (NFRL), independent of distribution assumptions, to capture historical driving information in latent space, significantly reducing sample complexity.
\item[•]We incorporate Safety Constraints (SC) directly into the reward function to enable the agent to account for safety considerations during training.
\item[•]By fine-tuning with a small amount of human expert dataset via using the G29 steering wheel, NFRL agents can learn more general driving principles, significantly improving generalization and sample efficiency.
\end{itemize}

\section{Related Work}
Imitation learning, which utilizes an efficient supervised learning approach, has gained widespread application in autonomous driving research due to its simplicity and effectiveness. For instance, imitation learning has been employed in end-to-end autonomous driving systems that directly generate control signals from raw sensor inputs \cite{LeMero2022ASO,codevilla2018end,bansal2018chauffeurnet,chen2019deep}.

Deep reinforcement learning (DRL) has demonstrated its strength in addressing complex decision-making and planning problems, leading to a series of breakthroughs in recent years. Researchers have been trying to apply deep RL techniques to the domain of autonomous driving.
Lillicarp et.al~\cite{lillicrap2015continuous} introduced a continuous control DRL algorithm that trains a deep neural network policy for autonomous driving on a simulated racing track. Wolf et.al~\cite{wolf2017learning} used Deep Q-Network to learn to steer an autonomous vehicle to keep in the track in simulation. Chen et.al~\cite{chen2018deep} developed a hierarchical DRL framework to handle driving scenarios with intricate decision-making processes, such as navigating traffic lights. Kendall et.al~\cite{kendall2019learning} marked the first application of DRL in real-world autonomous driving, where they trained a deep lane-keeping policy using only a single front-view camera image as input. Chen et.al~\cite{chen2021interpretable} proposed an interpretable DRL method for end-to-end autonomous driving. Nehme et.al~\cite{nehme2023safe} proposed safe navigation. Murdoch et.al~\cite{murdoch2023partial} propose a partial end-to-end algorithm that decouples the planning and control tasks. Zhou et.al~\cite{zhou2023identify} proposes a method to identify and protect unreliable decisions of a DRL driving policy. Zhang et.al~\cite{zhang2023spatial} a framework of constrained multi-agent reinforcement learning with a parallel safety shield for CAVs in challenging driving scenarios. Liu et.al~\cite{liu2022augmenting} propose the Scene-Rep Transformer to enhance RL decision-making capabilities. 

By combining the advantages of RL and IL is also a relatively popular method in recent years. The techniques outlined in \cite{theodorou2010reinforcement}, \cite{van2015learning}, \cite{zhou2023large} and \cite{zhao2023context} have proven to be efficient in merging demonstrations and RL for improving learning speed. Liu et.al~\cite{Liu2021ImprovedDR} propose a novel framework combining RL and expert demonstration to learn a motion control strategy for urban scenarios. Huang et.al~\cite{huang2023conditional} introduces a predictive behavior planning framework that learns to predict and evaluate from human driving data. Huang et.al~\cite{huang2024human} propose an enhanced human in-the-loop reinforcement learning method, while they rely on human expert performance and can only accomplish simple scenario tasks. DPAG \cite{rajeswaran2017learning} combines RL and imitation learning to solve complex dexterous manipulation problems. Our approach utilizes the potential for reinforcement learning and normalization flows to learn useful information from historical trajectory information, further learning expert demonstrations through DPAG methods.


\section{Methodology}

The proposed framework of our EGADS is illustrated in Figure~\ref{image1}. Firstly, human experts collect demonstrations offline using the G29 steering wheel. These expert demonstrations are then utilized as the RL fine-tuning experience replay buffers for training the entire model. Subsequently, a pre-training process is conducted to establish a model with human expert experience that does not update environmental data during training. The resulting model, enriched with human expert experience, is then used to fine-tune the policy for RL agent. Additionally, we have designed safety constraints for the intelligent vehicle, enhancing its safety performance. Furthermore, we explore 12 different types of images as inputs, which can be found in the Appendix ~\ref{sec:appendixA3}. 

\subsection{Preliminaries}
We model the control problem as a Partially Observable Markov Decision Process (POMDP), which is defined using the 7-tuple: \((S, A, T, R, \Omega, O, \gamma)\), where \(S\) is a set of states, \(A\) is a set of actions, \(T\) is a set of conditional transition probabilities between states, \(R\) is the reward function, \(\Omega\) is a set of observations, \(O\) is a set of conditional observation probabilities, and \(\gamma\) is the discount factor. The goal of the RL agent is to maximize expected cumulative reward $E[\sum_{t=0}^{\infty}\gamma_{t} r_{t}]$. After having taken action \(a_{t-1}\) and observing \(o_t\), an agent needs to update its belief state, which is defined as the probability distribution of the environment state conditioned on all historical information:
$b(s_t) = p(s_t \mid \tau_t, o_t)$, where \(\tau_t = \{o_1, a_1, \ldots, o_{t-1}, a_{t-1}\}\).

\begin{figure*}[!htp]
	\centering{\includegraphics[width=12cm]{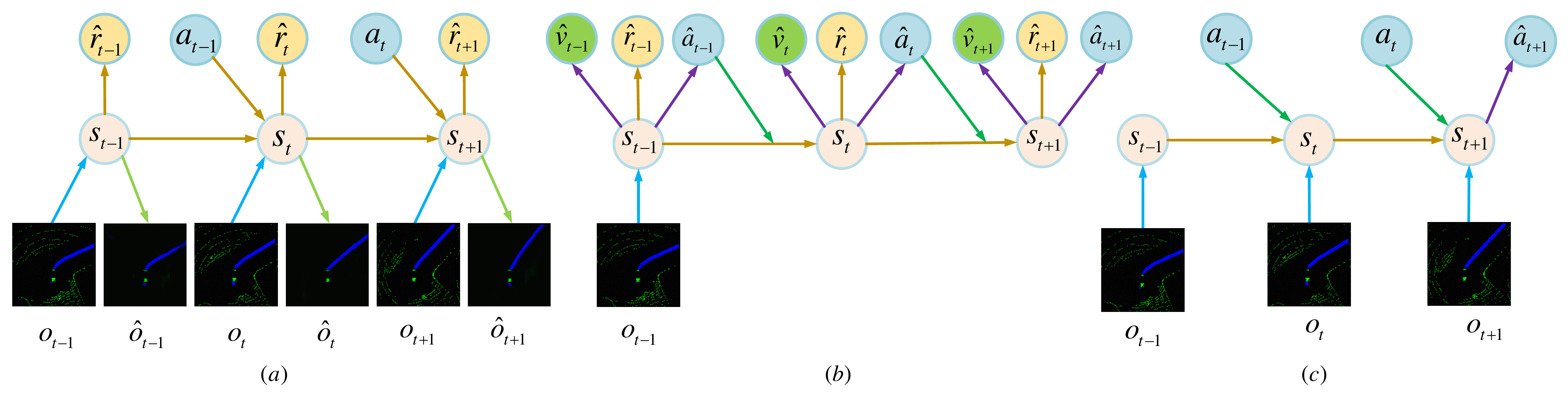}}
	\caption{(a) RL agent learns potential dynamics from past experience datasets. (b) RL agent predicts driving action in an imaginary space. (c) RL agent interacts with driving environment. Where $o$ is observation, $a$ is action, $s$ is latent state, $\hat{r}_{t}$ is reward, $\hat{o}_t$ is reconstructed and $\hat{v}_{t}$ is value. }
	\label{image2}
\end{figure*}
\subsection{Latent dynamic model for autonomous driving}
We propose the use of latent variables to solve problems in end-to-end autonomous driving.  This potential space is used to encode complex urban driving environments, including visual inputs, spatial features, and road conditions. Historical high-dimensional raw observation data is compressed into this low-dimensional latent space and learned through a sequential latent environment model that learns in conjunction with the maximum entropy RL process. 
We introduce RL agent model consists of components can be constructed the probabilistic graphical model of POMDP as follow:
\begin{equation}
\begin{aligned}
\text{State transition model:}&\quad p_{\theta}(s_{t}|s_{t-1},a_{t-1})\\
\text{Reward model:}&\quad p_{\theta}(r_{t}|s_{t})\\
\text{Observation model:}&\quad p_{\theta}(o_{t}|s_{t})\\
\end{aligned}
\end{equation}
where $p$ is prior probability, $q$ is posterior probability, $o$ is observation, $a$ is action,  is latent state and $\theta$ is the parameter of the model. Then the world model reconstructs the inputs images from the original sensors, more details can be found in the Appendix ~\ref{sec:appendixA7}.
\subsection{RL agent in the latent space}
Visual control \cite{thrun1999monte}, \cite{silver2010monte}, \cite{bengtsson2008curse} can be defined as a POMDP. The traditional components of agents that learn through imagination include dynamics learning, behavior learning, and environment interaction \cite{hafner2019dream}, \cite{hafner2019learning}. The RL agent in the latent space in our EGADS mainly includes the following:

(1) RL agent learns potential dynamics from past experience datasets of autonomous vehicle. As shown in Figure 2(a), using $p$ to represent prior probability, $q$ to represent posterior probability, agent learns to encode observation and action into compact latent state, and $\hat{o}_t$ is reconstructed with $q(\hat{o}_{t}|s_{t})$ while $s_t$ is determined via $p(s_{t}|s_{t-1}, a_{t-1}, o_{t})$.

(2) RL agent predicts driving action in an imaginary space. As shown in Figure 2(b), RL agent is in a close latent state space where it can predict value $\hat{v}_{t}$, reward $\hat{r}_{t}$ and action $\hat{a}_{t}$ based on current input $o_{t-1}$ with $q(\hat{v}_{t},\hat{r}_{t},\hat{a}_{t}|s_{t})$, $p(s_{t}|s_{t-1}, \hat{a}_{t-1})$, $q(\hat{a}_{t-1}|s_{t-1})$. 

(3) RL agent interacts with driving environment. As shown in Figure 2(c), RL agent predicts next action values $\hat{a}_{t+1}$ by encoding historical trajectory information via $q(\hat{a}_{t+1}|s_{t+1})$, $p(s_{t+1}|s_{t}, a_{t}, o_{t+1})$.

\subsection{Normalizing Flow for inferred belief }
Existing latent RL models in autonomous driving either suffer from the curse of dimensionality or make some assumptions and only learn approximate distributions. This approximation imposes strong limitations and is problematic, whereas distributions in the real world tend to be more flexible. In the continuous and dynamic space, existing methods based on normalizing flows (NF) \cite{dinh2016density} , \cite{huang2018neural}, \cite{rezende2015variational} can learn more flexible and generalized beliefs. These methods provide a solid foundation for RL agents to accurately predict future driving actions. Inspired by \cite{chen2022flow}, we added a belief inference model: $q_{\theta}(s_t|\tau_t,o_t)$, where $\theta$ is the parameter of the model. The belief model can be substituted for the probability density with NF in the KL-divergence term of equation 2.
\begin{small}
\begin{equation}
\begin{aligned}
&q_{K}(s_t|\tau_t,o_t)=\log q_{0}(s_t|\tau_t,o_t)-\sum_{k=1}^{K}\vert det \frac{\partial f_{\psi_k}}{\partial s_{t,k-1}}\vert\\
&p_{K}(s_t|\tau_t)=\log p_{0}(s_t|\tau_t)-\sum_{k=1}^{K}\vert det \frac{\partial f_{\omega_k}}{\partial s_{t,k-1}}\vert
\end{aligned}
\end{equation}
\end{small}
where $q_0=q_\theta$, $q_K = q_{\theta, \psi}$, $p_0=p_\theta$, $p_K=p_{\theta, \omega}$, $\psi$ and $\omega$ are the parameters of a series of mapping transformations of the posterior and prior distributions. Where $\tau_t=\{o_1, a_1,\cdots, o_{t-1}, a_{t-1} \}$. The input images $o_{1:t}$ and actions $a_{1:t-1}$ are encoded with $q_\theta(s_{t}|\tau_t, o_{t})$. Then the final inferred belief is obtained by propagating $q_{\theta}(s_t|\tau_t, o_t)$ through a set of NF mappings denoted $f_{\psi_K}\dots f_{\psi_1}$ to get a  posterior distribution $q_{\theta, \psi}(s_t|\tau_t, o_t)$. The final prior is obtained by propagating $p_{\theta}(s_t|\tau_t)$ through a set of NF mappings denoted $f_{\omega_K}\dots f_{\omega_1}$ to get a prior distribution $p_{\theta, \omega}(s_t|\tau_t)$. Where $p_{K}(s_t|\tau_t)=p_{K}(s_t|s_{t-1}, a_{t-1})$, given the sampled $s_{t-1}$ from $q_{K}(s_{1:t}|\tau_{t}, o_{t})$. Finally, our NF inference RL model (NFRL) is optimized by variational inference method, in which the evidence lower bound (ELBO) \cite{jordan1998introduction}, \cite{de2020block} is maximized. The loss function is defined as:
\begin{small}
\begin{equation}
\begin{aligned}
& \mathcal{M}_{\text {model}}(\theta,\psi,\omega)=\sum_{t=1}^{T}\big(\mathbb{E}_{q(s_{t} \mid o_{\leq t},a_{<t})}
[\log p_{\theta}(o_{t}\mid s_{t})+\\
&\log p_{\theta}(r_{t}\mid s_{t})]-\big.{\mathbb{E}_{q_K(s_{1:T}\mid o_{1:T},a_{1:T-1})}} [D_{\rm KL}\big(q_{K}(s_{t}\mid\tau_{t},o_{t})\|\\
&p_{K}(s_{t}\mid \tau_{t},o_{t}))]\big)
\end{aligned}
\end{equation}
\end{small}
\subsection{Policy optimization}
The action model implements the policy and is designed to predict the actions that are likely to be effective in responding to the simulated environment. The value model estimates the expected reward generated by the behavior model at each state $s_\tau$.
\begin{small}
\begin{equation}
\begin{aligned}
 &a_\tau \sim q_\phi(a_\tau|s_\tau),
 &v_{\eta}(s_\tau) = E_{q_\phi}[\sum_{t=t}^{t+H}\gamma^{\tau-t} r_{\tau}]
\end{aligned}
\end{equation}
\end{small}
where $\phi$, $\eta$ are the parameters of the approximated policy and value. The obejective of the action model is to use high value estimates to predict action that result in state trajectories
\begin{small}
\begin{equation}
\begin{aligned}
 &\mathcal{M}_{\text {actor}}(\phi) = E_{q_\phi}(\sum_{\tau=t}^{t+H}V_{\tau}^{\lambda})
\end{aligned}
\end{equation}
\end{small}
To update the action and value models, we calculate the value estimate $v_{\eta}(s_\tau)$ for all states $s_\tau$ along the imagined trajectory. $V_{\tau}^{\lambda}$ can be defined as follow:
\begin{small}
\begin{equation}
\begin{aligned}
&V_{\tau}^{\lambda}=(1-\tau)v_{\eta}(s_{\tau+1})+\lambda V_{\tau+1}^{\lambda},\quad\tau \textless {t+H}
\end{aligned}
\end{equation}
\end{small}
Then we can train the critic to regress the TD($\lambda$) \cite{sutton2018reinforcement} target return via a mean squared error loss:
\begin{small}
\begin{equation}
\begin{aligned}
\mathcal{M}_{\text {critic}}(\eta)=\mathbb{E}\Big[ \sum_{\tau=t}^{t+H} \frac{1}{2}\left(v_{\eta}\left(s_{\tau}\right)-V_{\tau}^{\lambda}\right)^{2}\Big]
\end{aligned}
\end{equation}
\end{small}
where $\eta$ denote the parameters of the critic network and $H$ is the prediction horizon. 
Then the loss function is as follows:
\begin{small}
\begin{equation}
\begin{aligned}
 \label{eq:2-6}
\underset{\psi,\eta,\phi,\theta,\omega,\eta}{\rm min} \alpha_0{\rm \mathcal M_{critic}(\eta)} -\alpha_1{\rm \mathcal M_{actor}(\phi)}-\alpha_2{\rm \mathcal M_{model}(\theta,\psi,\omega)}
\end{aligned}
\end{equation}
\end{small}
we jointly optimize the parameters of model loss $\psi, \theta, \omega$, critic loss $\eta$ and actor loss $\phi$, where $\alpha_0, \alpha_1, \alpha_2$ are coefficients for different components.

\subsection{Safety constraint}

In the Gym-Carla benchmark, the reward function proposed by Chen et.al \cite{chen2019model} is denoted as $f_1$. To ensure the intelligent vehicle operates safely and smoothly in complex environments, we incorporated additional safety and robustness constraints into $f_1$, denoted as $f_2=f_1+200r_{ft}+50r_{lt} + 2r_{sc}$. $r_{ft}$ is the front time to collision. $r_{lt}$ is lateral time to collision. $r_{sc}$ is the smoothness constraint. For detailed information on $f_1$ and $f_2$ of reward function, please refer to Appendix ~\ref{sec:appendixA4}.

(1) Front time to collision. When around vehicles are within the distance of ego vehicle (our agent vehicle) head in our setting, then we can calculate the front time to collision between ego vehicle and around vehicles. Firstly, the speed and steering vector $(s_{\tau},a_{\tau})\in \mathbb{S}$ of the ego vehicle are defined, where $s_{\tau}$ represents the angle vector of vehicle steering and $a_{\tau}$ represents the acceleration vector of the vehicle in local coordinate system. Secondly, two waypoints closest to the current ego vehicle are selected from the given navigation routing as direction vectors $w_p$ for the entire route progression, where $\rightarrow $ indicates a vector in world coordinates. The position vectors for both ego vehicle and around vehicles are represented by $(x_{t}^{*}, y_{t}^{*})$, respectively. Finally, $\delta_{e}$ and $\delta_{a}$ representing angles between position vectors for ego vehicle and around vehicles with respect to ${w_p}$ are calculated respectively.

\begin{small}
\begin{equation}
\begin{aligned}
 \label{eq:2-6}
 {w_p} =\big [(\frac{{w_{t+1}^{x}}-{w_{t}^{x}}}{2})-({w_{t}^{x}}),(\frac{{w_{t+1}^{y}}-{w_{t}^{y}}}{2})-({w_{t}^{y}})\big]\\
 \delta_{e} = \frac{[{v_{t}^{x*}},{v_{t}^{y*}}]\cdot {w_p}}{\Vert    {v_{t}^{x*}},{v_{t}^{y*}}\Vert_2\quad\Vert {w_p}\Vert_2},\delta_{a} = \frac{[{v_{t}^{x}},{v_{t}^{y}}]\cdot {w_p}}{\Vert {v_{t}^{x}},{v_{t}^{y}}\Vert_2\quad\Vert {w_p}\Vert_2}
\end{aligned}
\end{equation}
\end{small}
where, $l$ is the length of the set of waypoints $\mathbb{W}$ stored. The variable $t \in \tau$, and ${w_{t}^{x}}\in \mathbb{W}_1$ represents the $x$ coordinate of the first navigation point closest to the intelligent vehicle on its current route at time $t$. Similarly, ${w_{t+1}^{x}}\in \mathbb{W}_2$. Furthermore, it is possible to calculate the $F_{ttc}$ as follows:

\begin{small}
\begin{equation}
\begin{aligned}
 \label{eq:2-6}
 F_{ttc} = \frac{\Vert {x_{t}}-{x_{t}^{*}},{y_{t}}-{y_{t}^{*}}\Vert_2}{\left|\Vert {v_{t}^{x*}},{v_{t}^{y*}}\Vert_2sin(\delta_{e}) -\Vert {v_{t}^{x}},{v_{t}^{y}}\Vert_2sin(\delta_{a})\right|}
\end{aligned}
\end{equation}
\end{small}

(2) Lateral time to collision. When around vehicles are not within the distance of ego vehicle head in our setting, we consider significantly the $L_{ttc}$. The calculation method for $L_{ttc}$ and $F_{ttc}$ is the same. However, the collision constraint effect of $L_{ttc}$ on intelligent vehicle is limited, mainly due to the slow reaction time of intelligent vehicle to $L_{ttc}$, lack of robustness and generalization ability. Therefore, we have implemented a method of assigning values to different intervals for $L_{ttc}$ as follows:

\begin{small}
\begin{equation}
 \left\{
    \label{eq:2-5}
	\begin{aligned}
	&\text{min}( z_{\tau}, c_{\tau}+1.0),
	&\nu_{g}\leq (c_{\tau}-1.5) \text{ and } \mu_{a}\leq (c_{\tau}-0.5). \\
	&\text{min}( z_{\tau}, c_{\tau}-1.8),
	&\nu_{g}\leq (c_{\tau}-3.0) \text{ and } \mu_{a}\leq (c_{\tau}-2.0). \\
	&\text{min}( z_{\tau}, c_{\tau}-3.0),
	&\nu_{g}\leq (c_{\tau}-3.5) \text{ and } \mu_{a}\leq (c_{\tau}-3.0). \\
	\end{aligned}
	\right
	.
\end{equation}
\end{small}
where $c_{\tau}$ is the empirical const of $L_{ttc}$ in our setting at (5,7), $z_{\tau}$ is the ttc based on their combined speed. $\nu_{g}$ is the ttc obtained by calculating the longitudinal velocity. $\mu_{a}$ is the ttc obtained by calculating the lateral velocity.

(3) Smooth steering is defined as $\vert s_{t}^{\delta}-s_{t}^{*\delta}\vert \in e_{c}$. $s_{t}^{\delta}$ is the actual steering angle. $s_{t}^{*\delta}$ is the predicted steering angle based on policy $\pi$. The range of $e_{c}$ can be established based on empirical data.
\subsection{Augmenting RL policy with demonstrations}
Though, NFRL can significantly reduce complexity, and reward design based on safety constraints can enhance safety. Demonstrations can mitigate the need for painstaking reward shaping, guide exploration, further reduce sample complexity, and help generate robust, natural behaviors. We propose the demonstration augmented RL agent method which incorporates demonstrations into NFRL agent in two ways:

(1) Pretraining with behavior cloning. We use behavior cloning to provide a policy $\pi^*$ via expert demonstrations and then to train a model $\mathcal{M}_{\text{expert}}$ with some expert ability.
\begin{small}
\begin{equation}
 \label{eq:2-6}
\mathcal{M}_{\text{expert}} = {\underset{\xi}{\rm maximize}}\sum_{(s',a')\in \pi^{*}(\mathcal{D}_{e}) }\ln \pi^{*}_{\xi}(a_{\tau}'|s_{\tau}')
\end{equation}
\end{small}
where $\mathcal{D}_{e}$ is a human expert dataset obtained from driving G29 steering. For detailed information on all of the expert dataset, please refer to Appendix ~\ref{sec:appendixA2}.

(2) RL fine-tuning with augmented loss: we employ $\mathcal{M}_{\text{expert}}$ to initialize a model trained by deep RL policies, which reduces the sampling complexity of the deep RL policy. The training loss of the actor model as follows:
\begin{small}
\begin{equation}
\begin{aligned}
 \label{eq:2-6}
{\rm \mathcal M_{\hat{\text{actor}}}(\phi,\xi)}=\mathcal{M}_{\text {actor}}(\phi)+k \ln \pi^{*}_{\xi}(a_{\tau}'|s_{\tau}'), (a_{\tau}',s_{\tau}')\in \mathcal{D}_{e} 
\end{aligned}
\end{equation}
\end{small}
where $k$ represents the balance between the behavior cloning policy and NFRL policy, and is set as a constant based on empirical data. We only changed the actor model of NFRL, and the optimization of the other parts is exactly the same.

\begin{figure*}[!htp]
    \centering
    \includegraphics[width=1\linewidth]{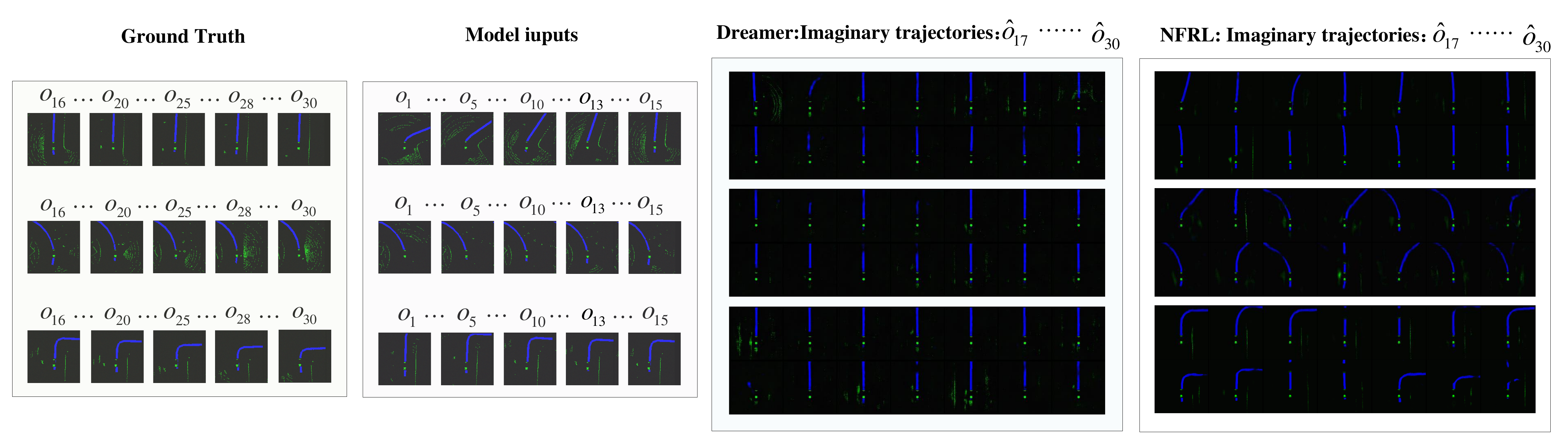}
    \caption{Randomly sample sensor inputs Lidar\_noground $o_1,o_2,\cdots,o_{15}$, and then our model can imagine driving behaviors $\hat{o}_{16},\hat{o}_{17},\cdots,\hat{o}_{30}$. The results show that, compared with Dreamer using ground truth data, our NFRL model is more accurate and diverse, with no mode mixup and less blur. }
    \label{fig:enter-label}
\end{figure*}

\section{Experiment}

\subsection{Experiment setup}

Models were trained on an NVIDIA RTX 3090 GPU using Python 3.8. Our experiments were conducted on a benchmark called Gym-carla, a third-party environment for OpenAI gym that is used with the CARLA simulator\cite{dosovitskiy2017carla}. 
In our experiments, the NFRL series models and baseline methods were trained in Town03 (random) and evaluated across four scenarios: Town03, Town04, Town05, and Town06. These scenarios are abbreviated as Town03-Town06, encompassing both random and roundabout modes. Town03 simulates a realistic urban environment with diverse features such as tunnels, intersections, roundabouts, curves, and turnaround bends. Descriptions of other maps are provided in Appendix ~\ref{sec:appendixA1}. Here, "random" refers to randomly selected intersections and driving scenarios, while "roundabout" focuses specifically on roundabout intersections.
We present the hyperparameter settings for the methods in Appendix ~\ref{sec:appendixA6}. Particularly we contribute a dataset of human expert $\mathcal{D}_{expert}$ via G29 steering wheel, with further details provided in Appendix ~\ref{sec:appendixA2} description. 

\subsection{Measure Driving Performance Metrics}
In the Gym-Carla benchmark, an episode terminates under any of the following conditions: the number of collisions exceeds one, the maximum number of time steps is reached, the destination is reached, the cumulative lateral deviation from the lane exceeds 10 meters, or the vehicle remains stationary for 50 seconds. EGADS is an end-to-end autonomous driving system. We implemented our trained model on an autonomous vehicle for urban navigation, assessing performance through five standard metrics:  Off-road Rate (OR), Episode Completion Rate (ER), Average Safe Driving Distance (ASD), Average Reward (AR) using the reward function from Chen et.al \cite{chen2019model} that accounts for driving dynamics (yaw, collisions, speeding, and lateral velocity), and Driving Score (DS) - a composite metric calculated as DS = ER × AR in accordance with CARLA Leaderboard standards. During model selection, we focused on checkpoints that simultaneously optimized DS and AR, while implementing the remaining metrics (ER, OR, AR, ASD) based on the methodology from Gao et.al \cite{gao2024enhance} and Tang et.al \cite{tangvlascd, tang2024efficient}. For detailed information on all of the above metrics, please refer to Appendix ~\ref{sec:appendixA5}. 

\subsection{Comparison settings}
In order to evaluate the performance of our autonomous driving system more effectively, we have conducted various comparisons with existing methods such as DDPG \cite{lillicrap2015continuous}, SAC \cite{haarnoja2018soft}, TD3 \cite{fujimoto2018addressing}, DQN \cite{mnih2015human}, Latent\_SAC \cite{chen2021interpretable}, Dreamer \cite{hafner2019dream}, CQL \cite{kumar2020conservative}. We decomposed EGADS into three components: NFRL, SC (safety constraint) and augmenting RL policy with demonstrations. We decomposed EGADS into three components: NFRL, Safety Constraint (SC) and augmenting RL policy with demonstrations (Demo). We then conducted evaluations using four comparison settings, NFRL, NFRL+SC, BC+Demo, NFRL+SC+Demo. BC+Demo indicates the use of behavioral cloning to imitate the expert dataset, while NFRL+SC+Demo involves using expert datasets to augment the NFRL policy combined with SC.

\begin{table}[]
\caption{In training, all methods were compared under different RL baselines in Town03 (random), with episodes of 500 steps. $+\infty$ indicates failure to reach the baseline within the maximum tested runtime of 250 GPU hours. }
\label{sample-table}
\centering
\resizebox{0.5\textwidth}{!}{%
\begin{tabular}{ccccc}
\toprule
Method & \multicolumn{2}{c}{ASD=50m}  &\multicolumn{2}{c}{ASD=100m} \\
\cmidrule(lr){2-5}
&episodes$\downarrow$&times$\downarrow$&episodes$\downarrow$&times $\downarrow$ \\
\midrule
DDPG &$+\infty$&$+\infty$&$+\infty$&$+\infty$ \\
SAC  &$+\infty$ &$+\infty$ &$+\infty$ &$+\infty$\\
TD3 &$\geq$161&$\geq$192h &$+\infty$&$+\infty$\\
DQN &$\geq$163&$\geq$53h &$+\infty$&$+\infty$  \\
Latent\_SAC  &$\geq$167&$\geq$43h &$\geq$352&$\geq$105h   \\
Dreamer &$+\infty$&$+\infty$&$+\infty$&$+\infty$  \\
NFRL(our) &$\geq$141 &\textbf{$\geq$21h}  &$\geq$121 & \textbf{$\geq$65h}\\
\bottomrule
\end{tabular}%
}
\end{table}

\begin{table*}[]
\caption{In training, all methods were compared under different NFRL baselines in Town03 (random), with episodes of 500 steps. $+\infty$ indicates failure to reach the baseline within the maximum tested runtime of 250 GPU hours. }
\label{sample-table}
\centering
\resizebox{0.7\textwidth}{!}
{%
\begin{tabular}{ccccccc}
\toprule
Method & \multicolumn{2}{c}{ASD=50m}  &\multicolumn{2}{c}{ASD=100m}  &\multicolumn{2}{c}{ASD=200m} \\
\cmidrule(lr){2-7}
&episodes$\downarrow$&times$\downarrow$&episodes$\downarrow$&times $\downarrow$ &episodes$\downarrow$&times $\downarrow$ \\
\midrule
NFRL &$\geq$141 &\textbf{$\geq$21h}  &$\geq$121&\textbf{$\geq$65h }&$+\infty$&$+\infty$\\
NFRL+SC  &$\geq$71 &$\geq$12h &$\geq$301&$\geq$40h&$\geq$1100 &$\geq$146h\\
NFRL+SC+Demo  &$\geq$21 &\textbf{$\geq$1.3h} &$\geq$58&\textbf{$\geq$3h} &$\geq$321&\textbf{$\geq$48h}\\
\bottomrule
\end{tabular}%
}
\end{table*}
\subsection{Results on trajectory prediction}
In order to accurately evaluate our model prediction of driving actions for intelligent vehicle, this problem can be viewed as a special POMDP problem with the reward value maintained at 0. As shown in Figure~\ref{fig:enter-label}, the comparison with ground-truth data demonstrates that our NFRL model achieves higher accuracy and greater diversity than Dreamer, with no mode collapse and significantly reduced blurring effects. We provide additional results on predictions of future driving actions for NFRL in Appendix ~\ref{sec:appendixA8}.

\subsection{How to reduce sampling complexity ?}
To evaluate the sampling complexity of different methods, we used the average ASD  as the test threshold and set three distinct checkpoints at 50m, 100m, and 200m. We measured the GPU hours required for each method to reach the corresponding ASD threshold, with a maximum testing duration capped at 250 GPU hours, as shown in Tables 1 and 2. Notably, in Table 1, although different methods require varying numbers of episodes to reach the ASD threshold, the actual time consumed differs significantly. This is because each episode has a fixed length of 500 steps. Some methods remain stationary for most of the episode, yet the episode does not terminate early, leading to prolonged total runtime. In contrast, other methods may collide or deviate from the lane, triggering early termination of the episode. 

As shown in Table 1, our proposed NFRL method significantly improves training time efficiency, achieving at least a 2-fold acceleration in reaching the 50-meter and 100-meter baselines compared to existing reinforcement learning methods. However, due to frequent collision issues observed in experiments, the method fails to surpass the 150-meter baseline. To address this limitation, we innovatively design a reward function incorporating Safety Constraints (SC). Experimental results, as presented in Table 2, show that the enhanced NFRL+SC method not only successfully achieves the 200-meter baseline but also improves training efficiency by at least 1.5 times compared to the original NFRL method. To further optimize performance, we introduce expert datasets for fine-tuning. Experimental data indicate that the NFRL+SC+Demo method achieves a remarkable 3-fold improvement in training efficiency over the NFRL+SC method when reaching the 200-meter baseline. 

The performance improvements are primarily driven by three key mechanisms: (1) The NFRL framework employs Normalizing Flow technology to reconstruct training data distributions, aligning them more closely with real-world driving scenarios. This technique enables both accurate future trajectory prediction and comprehensive coverage of possible trajectories across diverse driving situations. Such high-quality data representation allows the model to rapidly learn correct behavioral patterns. (2) The Safety Constraint (SC) module dynamically limits the policy exploration scope to safe regions, thereby minimizing costly divergent behaviors. (3) Demonstration data accelerates reward function discovery by injecting domain-specific prior knowledge. This co-design enables EGADS to achieve efficient convergence in complex autonomous driving scenarios, establishing it as a paradigm for sample-efficient reinforcement learning.

\begin{table*}[t]
\centering
\caption{Performance Comparison Across multiple Towns (Trained in Town03, Evaluated in Town04-Town06, hereinafter referred to as T04-T06)}
\label{tab:final_without_t01}
\resizebox{\textwidth}{!}{%
\small  
\begin{tabular}{@{}l|ccc|ccc|ccc|ccc|ccc@{}}
\toprule
\multirow{2}{*}{Method} & 
\multicolumn{3}{c}{DS $\uparrow$} & 
\multicolumn{3}{c}{AR ($f_1$) $\uparrow$} & 
\multicolumn{3}{c}{EC (\%) $\uparrow$} & 
\multicolumn{3}{c}{OR (\%) $\downarrow$} & 
\multicolumn{3}{c}{ASD (m) $\uparrow$} \\
\cmidrule(lr){2-4} \cmidrule(lr){5-7} \cmidrule(lr){8-10} \cmidrule(lr){11-13} \cmidrule(lr){14-16}
& T04 & T05 & T06 & T04 & T05 & T06 & T04 & T05 & T06 & T04 & T05 & T06 & T04 & T05 & T06 \\
\midrule
DDPG       & -0.10 & -0.01 & -0.08 & -10.01 & -10.1 & -10.02 & 0.00 & 0.01 & 0.00 & - & - & - & 0.00 & 0.00 & 0.00 \\
DQN        & 17.50 & 60.67 & 69.09 & 76.37 & 174.89 & 206.66 & 11.38 & 15.84 & 16.34 & 11.83 & 11.83 & 15.26 & 20.29 & 31.01 & 36.83 \\
TD3        & -15.89 & -2.24 & -25.62 & -131.36 & -84.30 & -195.60 & 9.18 & 8.16 & 5.82 & 33.32 & 16.94 & 16.02 & 6.17 & 10.05 & 4.50 \\
SAC        & -20.56 & -14.89 & -15.02 & -14.08 & -18.92 & -16.67 & 4.95 & \textbf{69.07} & \textbf{85.60} & 0.00 & 0.00 & 0.00 & 6.71 & 6.07 & 8.03 \\
L\_SAC     & 102.61 & 110.66 & 21.52 & 170.79 & 8.70 & 145.77 & 15.09 & 12.78 & 12.21 & 1.05 & 4.96 & 4.64 & 15.97 & 21.24 & 42.15 \\
Dreamer    & -0.01 & -0.03 & -0.03 & -15.10 & -15.10 & -15.20 & 0.00 & 0.12 & 0.20 & - & - & - & 0.01 & 0.01 & 0.00 \\
NFRL (base)& \textbf{326.78} & \textbf{390.54} & \textbf{431.44} & \textbf{1509.90} & \textbf{785.92} & \textbf{947.26} & \textbf{15.81} & 22.61 & 29.59 & 30.88 & 12.05 & 16.50 & \textbf{220.18} & \textbf{123.24} & \textbf{143.61} \\
\bottomrule
\end{tabular}%
}
\end{table*}

\begin{table*}[t]
\centering
\caption{Evaluation results for different methods in CARLA Town03 (random) and Town03 (roundabout): we denote RND as random and RBT as roundabout. For a fair comparison, all reward functions are in the form of $f_1$. Particularly, $-$ indicates that valid data could not be obtained because the episode completion rate for this method is close to 0.}
\resizebox{0.8\textwidth}{!}{%
\small
\begin{tabular}{ccccc|cccccc}
\toprule
Method & \multicolumn{2}{c}{DS $\uparrow$} & \multicolumn{2}{c}{AR ($f_1$) $\uparrow$} & \multicolumn{2}{c}{EC(\%) $\uparrow$} & \multicolumn{2}{c}{OR(\%) $\downarrow$} & \multicolumn{2}{c}{ASD(m) $\uparrow$} \\
\cmidrule(lr){2-3} \cmidrule(lr){4-5} \cmidrule(lr){6-7} \cmidrule(lr){8-9} \cmidrule(lr){10-11}
& RND & RBT & RND & RBT & RND & RBT & RND & RBT & RND & RBT \\
\midrule
DDPG       & $-0.11$ & $-0.08$ & $-10.01$ & $-10.02$ & $0.01$ & $0.00$ & $-$ & $-$ & $0.00$ & $0.00$ \\
DQN        & $30.64$ & $36.33$ & $86.50$ & $121.24$ & $17.02$ & $16.42$ & $8.52$ & $11.83$ & $21.68$ & $26.27$ \\
TD3        & $2.40$ & $-6.60$ & $-18.15$ & $-129.52$ & $6.91$ & $4.07$ & $51.53$ & $49.32$ & $7.51$ & $3.12$ \\
SAC        & $-7.57$ & $-20.56$ & $-19.90$ & $-24.74$ & $\textbf{63.76}$ & $\textbf{67.95}$ & $\textbf{0.00}$ & $\textbf{0.65}$ & $6.27$ & $6.71$ \\
L\_SAC     & $125.95$ & $31.59$ & $161.24$ & $84.13$ & $11.98$ & $10.50$ & $14.72$ & $1.14$ & $31.31$ & $13.87$ \\
Dreamer    & $-0.03$ & $-0.02$ & $-15.12$ & $-15.12$ & $0.01$ & $0.02$ & $-$ & $-$ & $0.01$ & $0.01$ \\
NFRL (base)& $\textbf{170.03}$ & $\textbf{48.73}$ & $\textbf{424.60}$ & $\textbf{249.17}$ & $24.04$ & $10.88$ & $20.93$ & $18.11$ & $\textbf{72.16}$ & $\textbf{46.27}$ \\
\bottomrule
\end{tabular}%
}
\end{table*}

\subsection{Ablation study}
In the ablation study of the EGADS system, we evaluated the contributions of each module in cross-domain scenarios (evaluated in Town04, Town05 and Town06, trained in Town03) to validate the generalization performance of the NFRL, SC and Demo modules, as shown in Tables 5. The driving score (DS) served as the primary comprehensive metric, with other indicators providing supplementary reference. The addition of the SC module significantly improves the cross-scenario performance of NFRL (e.g. NFRL vs. NFRL + SC), demonstrating the effectiveness of our SC module design. Further incorporating the Demo learning module on top of NFRL+SC, the experimental results show that NFRL+SC+Demo achieves the highest scores in Town04 (1174.16), Town05 (723.90), and Town06 (2155.40), with substantial improvements over both the baseline NFRL and NFRL+SC configurations. This proves that the Demo module enhances cross-domain generalization through expert knowledge.

As shown in Table 6, we conducted comprehensive comparisons with various mainstream baselines (online RL methods such as L\_SAC and Dreamer; offline or imitation learning approaches including BC+Demo and CQL+Demo) across two challenging scenarios (Town03 RND and RBT). The multi-dimensional evaluation metrics clearly demonstrate that: 1) The NFRL framework itself surpasses existing online RL methods; 2) The SC module universally and significantly enhances both safety and overall performance across all methods, including baselines; 3) The NFRL framework effectively utilizes demonstration data, achieving far superior results compared to imitation learning and offline RL baselines; 4) The final NFRL+SC+Demo solution comprehensively outperforms all methods, including enhanced baselines, across nearly all positive metrics (DS, AR, EC, ASD) while maintaining excellent safety performance. These results fully validate the absolute superiority of our proposed method, the effectiveness of each module, and the powerful synergistic effects of their combination.

\begin{table*}[t]
\centering
\caption{During the evaluation, an ablation study of EGADS's three modules across scenarios was conducted (Trained in Town03, Evaluated in Town04-Town06, hereinafter referred to as T04-T06)}
\label{tab:ablation_study}
\resizebox{\textwidth}{!}{%
\small  
\begin{tabular}{@{}l|ccc|ccc|ccc|ccc|ccc@{}}
\toprule
\multirow{2}{*}{Method} & 
\multicolumn{3}{c}{DS $\uparrow$} & 
\multicolumn{3}{c}{AR ($f_1$) $\uparrow$} & 
\multicolumn{3}{c}{EC (\%) $\uparrow$} & 
\multicolumn{3}{c}{OR (\%) $\downarrow$} & 
\multicolumn{3}{c}{ASD (m) $\uparrow$} \\
\cmidrule(lr){2-4} \cmidrule(lr){5-7} \cmidrule(lr){8-10} \cmidrule(lr){11-13} \cmidrule(lr){14-16}
& T04 & T05 & T06 & T04 & T05 & T06 & T04 & T05 & T06 & T04 & T05 & T06 & T04 & T05 & T06 \\
\midrule
NFRL   & 326.78 & 390.54 & 431.44 & \textbf{1509.90} & 785.92 & 947.26 & 15.81 & 22.61 & 29.59 & 30.88 & 12.05 & 16.50 & \textbf{220.18} & \textbf{123.24} & 143.61 \\
NFRL+SC       & 649.46 & 213.17 & 1234.89 & 418.22 & 381.39 & 1571.85 & 48.70 & 38.80 & 63.42 & \textbf{0.00} & \textbf{0.00} & \textbf{0.00} & 91.34 & 50.42 & 195.36 \\
NFRL+SC+Demo  & \textbf{1174.16} & \textbf{723.90} & \textbf{2155.40} & 1329.89 & \textbf{894.58} & \textbf{2294.30} & \textbf{57.41} & \textbf{46.85} & \textbf{82.47} & 3.25 & 11.51 & 6.05 & 159.42 & 116.96 & \textbf{265.92} \\
\bottomrule
\end{tabular}%
}
\end{table*}

\subsection{How to improve generalization capabilities ?}
\begin{table*}[t]
\centering
\caption{Evaluation results for different methods in CARLA Town03 (random) and Town03 (roundabout): we denote RND as random and RBT as roundabout. For a fair comparison, all reward functions are in the form of $f_1$. Particularly, $-$ indicates that valid data could not be obtained because the episode completion rate for this method is close to 0.}
\resizebox{0.8\textwidth}{!}{%
\begin{tabular}{ccc|cc|cccccc}
\toprule
Method & \multicolumn{2}{c}{DS $\uparrow$} & \multicolumn{2}{c}{AR ($f_1$) $\uparrow$} & \multicolumn{2}{c}{EC(\%) $\uparrow$} & \multicolumn{2}{c}{OR(\%) $\downarrow$} & \multicolumn{2}{c}{ASD(m) $\uparrow$} \\
\cmidrule(lr){2-3} \cmidrule(lr){4-5} \cmidrule(lr){6-7} \cmidrule(lr){8-9} \cmidrule(lr){10-11}
& RND & RBT & RND & RBT & RND & RBT & RND & RBT & RND & RBT \\
\midrule
L\_SAC     & $125.95$ & $31.59$ & $161.24$ & $84.13$ & $11.98$ & $10.50$ & $14.72$ & $1.14$ & $31.31$ & $13.87$ \\
Dreamer    & $-0.03$ & $-0.02$ & $-15.12$ & $-15.12$ & $0.01$ & $0.02$ & $-$ & $-$ & $0.01$ & $0.01$ \\
NFRL    & 170.03 & 48.73  & 424.60 & 249.17 & 24.04 & 10.88 & 20.93 & 18.11 & 72.16 & 46.27 \\
\midrule
L\_SAC+SC     & $156.23$ & $64.76$ & $284.02$ & $148.50$ & $13.98$ & $15.91$ & $10.64$ & $12.88$ & $50.52$ & $18.67$ \\
Dreamer+SC    & $98.12$ & $50.02$ & $124.74$ & $74.98$ & $10.42$ & $11.20$ & $18.08$ & $16.35$ & $42.85$ & $16.90$ \\
NFRL+SC       & 192.84 & 101.29 & 341.56 & 181.28 & 38.46 & 34.66 & \textbf{5.87}  & \textbf{4.04}  & 80.21 & 50.24 \\
\midrule
BC+Demo & $-6.30$ & $-1.63$ & $-62.43$ & $-27.92$ & $9.22$ & $10.31$ & $15.49$ & $15.57$  & $14.78$ & $15.34$ \\
CQL+Demo & $8.52$ & $4.35$ & $42.10$ & $49.06$ & $10.58$ & $8.21$ & $13.45$ & $19.08$  & $19.25$ & $16.01$ \\
NFRL+Demo & $203.26$ & $143.03$ & $478.04$ & $26.15$ & $25.71$ & $20.66$ & $10.50$ & $12.82$  & $81.32$ & $64.80$ \\
NFRL+SC+Demo  & \textbf{485.92} & \textbf{380.17} & \textbf{720.27} & \textbf{653.21} & \textbf{44.25} & \textbf{36.63} & 7.69  & 5.48  & \textbf{100.13} & \textbf{84.92} \\
\bottomrule
\end{tabular}%
}
\end{table*}
The EGADS system enhances cross-scenario generalization through the co-design of the NFRL framework, SC module, and Demo module. NFRL decouples state representation from policy learning, establishing a transferable foundation for driving policies. As shown in Table 3, in the cross-town evaluation (Town04-Town06), NFRL achieves a significantly higher DS value compared to traditional reinforcement learning methods, demonstrating robust generalization capabilities.  

As evidenced in Tables 5 and 6, the SC module effectively mitigates high-risk behaviors through trajectory smoothing, improving overall DS values compared to standalone NFRL and enhancing system robustness. Meanwhile, the Demo module accelerates policy convergence and optimizes exploration via imitation learning. As shown in Tables 5 and 6, the NFRL+SC+Demo configuration demonstrates significant improvements across multiple metrics including DS and AR , confirming that demonstration data effectively reduces inefficient sampling.

The synergy between the SC module and Demo data can be summarized as follows: the SC module establishes safety boundaries to prevent the policy from entering hazardous or suboptimal states, while Demo data alleviates the conservatism of the SC module. EGADS integrates imitation learning (BC loss) and reinforcement learning (NFRL loss), dynamically balancing their weights to enable the agent to leverage expert knowledge while exploring autonomously within safe limits. This balanced mechanism enhances the policy's generalization capability and environmental adaptability, enabling efficient task execution across diverse scenarios and rapid adaptation to new challenges.

\section{Conclusion}
In summary, our EGADS framework effectively enhances sample efficiency, safety, and generalization in autonomous driving systems. The inclusion of safety constraints significantly enhances vehicle safety. NFRL, our proposed method, accurately predicts future driving actions, reducing sample complexity. By fine-tuning with a small amount of expert data, NFRL agents learn more general driving principles, which greatly improve generalization and sample complexity reduction, offering valuable insights for autonomous driving system design.
\bibliography{neurips_2024}
\bibliographystyle{plain}
\clearpage

\appendix

\section{Appendix}
In the appendix, we provide more details regarding the efficient and generalized end-to-end autonomous driving system with latent deep reinforcement Learning and demonstrations in the paper, including
\begin{itemize}
\item[•]In Subsection A.1, we provide a detailed description of the maps used for training in Gym-Carla.
\item[•]In Subsection A.2, we explain how we collected the expert demonstration dataset in CARLA using a human expert driving with the G29 steering wheel. 
\item[•]In Subsection A.3, we explore the impact of up to 12 different data input types on the performance of the NFRL agent. 
\item[•]In Subsection A.4, we introduce our reward function with safety constraints.
\item[•]In Subsection A.5, we provide a comprehensive measurement of driving performance metrics.
\item[•]In Subsection A.6, we present the hyperparameter settings for the methods involved in our experiments.
\item[•]In Subsection A.7, we demonstrate the reconstruction of original sensor input images by our NFRL
\item[•]In Subsection A.8, we provide additional results on predictions of future driving actions for NFRL in the imagination space.

\end{itemize}

\subsection{Training CARLA maps}
\label{sec:appendixA1}
In order to comprehensively evaluate the performance of our EGADS, we utilized four maps in CARLA, Town03, Town04, Town05 and Town06 as shown in Figure 4. Town04, a small town embedded in the mountains with a special infinite highway. Town05, squared-grid town with cross junctions and a bridge. Town06, long many lane highways with many highway entrances and exits. Particularly, Town03 is the most complex town with a 5-lane junction, a roundabout, unevenness, a tunnel, and more.

\subsection{Collect expert datasets}
\label{sec:appendixA2}
CARLA can be operated and controlled through using the python API. Figure 5 shows that we establish a connection between the Logitech G29 steering wheel and the CARLA, and then human expert can collect the datasets of teaching via the G29 steering wheel. Specifically, we linearly map accelerator pedals, brake pedals, and turning angles into \emph{accel[0,3](min,max)}, \emph{brake[-8,0] (min,max)}, \emph{steer[-1,1](left,right)}. The tensors are written into user-built Python scripts and combined with CARLA built-in Python API so that users can provide input from their steering wheels to autonomous driving cars in CARLA simulator for $\mathcal{D}_{expert}$ collection. Particularly, we contributed a dataset collected through human expert steering wheel control.

\subsection{Multiple types of input images}
\begin{figure*}
	\centering
    \includegraphics[width=12cm]{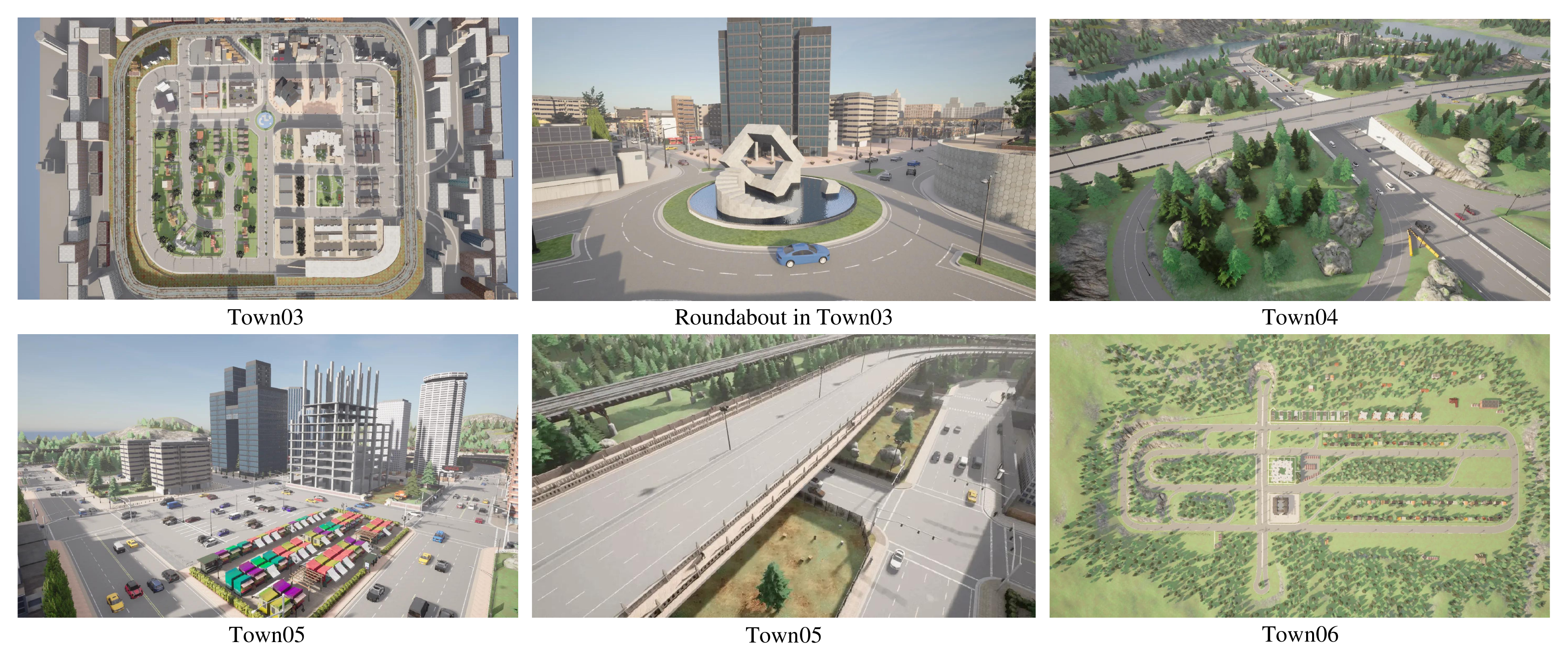}
    \caption{The road networks of the CARLA include routes for  Town03, Town04, Town05, and Town06}
\end{figure*}
\begin{figure*}[htbp]
	\centering
	\includegraphics[width=12cm]{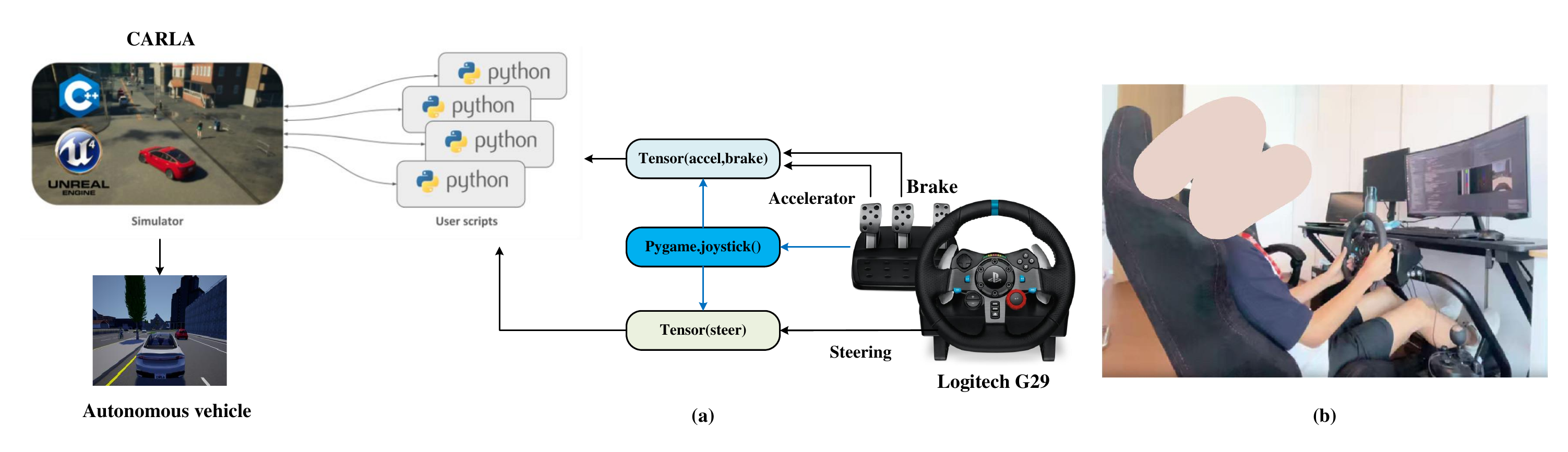}
	\caption{(a) CARLA connects with the G29 steering wheel (b) Human expert collects the datasets via the G29 steering wheel}
	\label{fig.2-4}
\end{figure*}

\label{sec:appendixA3}
The 12 types of input data we designed are mainly categorized into single-modal and single-image input, single-image and multimodal fusion, and multiple images and multimodal fusion, as shown in Figure 6. We compare various input image types for evaluating the performance of NFRL, as shown in Table 7. ASD of lidar\_noground reaches the highest value compared with all other input types. This is because lidar\_noground removes a large amount of redundant information, reduces the difficulty of world model understanding environment semantics, and also involves stationary status of intelligent vehicle in experiment. The results show that the lidar\_noground input is relatively optimal. However, it is worth noting that the effects of these 12 different input types are relatively small, with the ASD only varying between 20 and 40 meters. This shows that different data types have a minimal impact on the safety performance of intelligent vehicles.

\begin{table*}[h]
\caption{Evaluate the performance of NFRL under multiple input images in town03, training steps=100k. ASD is based on 5 episodes, the number of vehicles is 200.}
\label{image}
\centering
\resizebox{50mm}{!}{%
\begin{tabular}{lcc}
\toprule
Multiple input & ASD (m) \\
\midrule
birdeye & 29.4 \\
lidar & 38.6 \\
camera & 41.5 \\
lidar + camera & 56.4 \\
semantic & 26.3 \\
depth & 29.4 \\
lidar\_ng & \textbf{64.7} \\
multi-fusion1 & 52.1 \\
multi-fusion2 & 47.2 \\
lidar + depth\_ng & 32.5 \\
lidar\_ng + multi-fusion3 & 36.4 \\
lidar\_ng + camera\_gray & 48.8 \\
\bottomrule
\end{tabular}%
}
\end{table*}

1)\textbf{Single-modal and input of a single image}.  As Figure 6 shown, the lidar images, which project the 3D point cloud information from lidar onto a 2D point cloud image, with each pixel color determined by whether there is lidar or other relevant pixel information on the corresponding area. Navigation path is rendered in blue and surrounding road conditions are represented by green rectangular boxes to indicate participants such as vehicles, pedestrians etc. Particularly, lidar\_noground is created to remove redundant ground truth information from the 2D point cloud image. Moreover, we also consider camera, semantic, birdeye and depth as our sensor inputs. 

2)\textbf{Single-image and multimodal fusion}. The input of single-image and multi-modal fusion involve fusing lidar, rgb forward-facing grayscale image (camera\_gray), and navigation path into a composite rgb image with three types of information. The fused image has three channels, multi-fusion1 (lidar, camera\_gray, routing). Similarly having multi-fusion2 (lidar,depth,routing) and multi-fusion3 (lidar, depth,0).

3)\textbf{Multiple images and multi-modal fusion}. Multiple fusion can complement the shortcomings of a single input source and provide richer and more effective information. Therefore, we also design several single-modal fusion inputs as shown on the right side of Figure 6, including lidar\_noground and multi\_fusion3, lidar\_noground and depth, lidar-noground and camera\_gray, as well as camera and lidar.
\begin{figure*}[htbp]
	\centering
	\includegraphics[width=12cm]{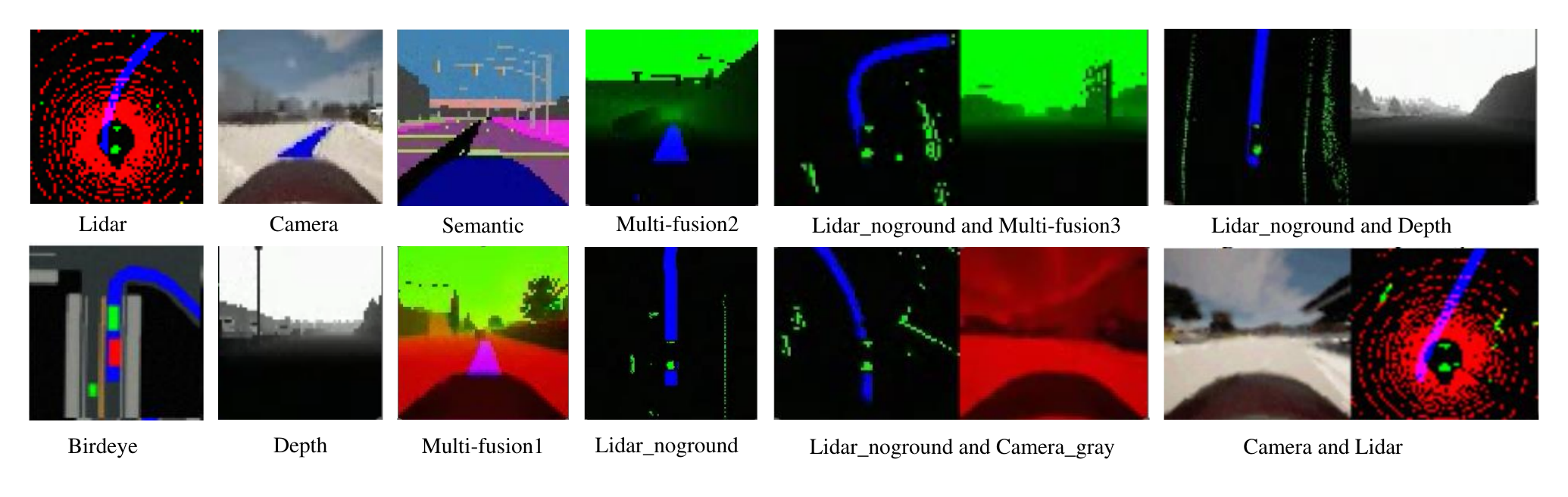}
	\caption{Multiple types of input images}
	\label{fig.2-4}
\end{figure*}

\subsection{Reward function}
\label{sec:appendixA4}
We use the following reward function $f_2$ in our experiments:
where $r_c$ is the reward related to collision, which is set to -1 if the ego vehicle collides and $0$ otherwise. $v_{lon}$ is the longitudinal speed of the ego vehicle. $r_f$ is the reward related to running too fast, which is set to $-1$ if it exceeds the desired speed (8 m/s here) and $0$ otherwise. $r_o$ is set to $-1$ if the ego vehicle runs out of the lane, and $0$ otherwise. $\alpha$ is the steering angle of the ego vehicle in radians. $r_{lat}$ is the reward related to lateral acceleration, which is calculated by $r_{lat} = -|\alpha| \cdot v_{lon}^2$. The last constant term is added to prevent the ego vehicle from standing still. $r_{ft}$ represents the time to collision in the forward direction, and if it is an autonomous vehicle and the time to collision with surrounding vehicles is below the safety threshold, this term is set to -1. $r_{lt}$ represents the time to collision in the lateral direction, and if it is an autonomous vehicle and the time to collision with surrounding vehicles is below the safety threshold, this term is set to -1. $r_{sc}$ represents the smoothness constraint, and if the actual steering angle of the autonomous vehicle differs significantly from the predicted steering angle by the model, exceeding a set empirical constant, this term is set to -1.
\begin{equation}
\begin{aligned}
\label{eq:2-6}
&f_1 =200r_c+ v_{lon} + 10r_f + r_o - 5 \alpha^2 + 0.2r_{lat} - 0.1\\
&f_2=f_1+200r_{ft}+50r_{lt} + 2r_{sc} 
\end{aligned}
\end{equation}
where the reward function $f_1$ is proposed by \cite{chen2019model}.

\subsection{Measure performance metrics}
\label{sec:appendixA5}
We use multiple key metrics to evaluate the performance of autonomous driving models in various driving scenarios. Outlane Rate (OR): the rate at which the vehicle deviates from its designated lane.  This metric evaluates the ability of modes to maintain proper lane discipline. Episode Completion Rate (ER): the percentage of driving tasks or episodes that the vehicle successfully completes. Higher completion rates indicate better task performance. Average Safe Driving Distance (ASD): the average distance driven without incidents, such as collisions or off-road events.  This metric highlights the capability to drive safely over extended periods. Average Return (AR): A metric that measures the cumulative reward collected by the vehicle during its driving tasks, often reflecting both task performance and adherence to safety guidelines. Driving Score (DS): A comprehensive metric that reflects the overall performance of the vehicle in terms of safety, efficiency, and compliance with traffic rules.
\begin{equation}
 OR = \frac{N_{\text{off\_road\_events}}}{N_{\text{total\_episodes}}}, ER = \frac{N_{\text{completed\_steps}}}{N_{\text{total\_steps}}}, AR = \frac{\sum_{i=1}^{N_{\text{episodes}}} \text{rewards}_i}{N_{\text{total\_episodes}}}
\end{equation}

\begin{equation}
ASD = \frac{\sum_{i=1}^{N_{\text{episodes}}} \text{distance}_i}{N_{\text{total\_episodes}}}, DS = ER \times AR  
\end{equation}
Where \(N_{\text{total\_episodes}}\) is the total number of episodes in the test. Where \(N_{\text{off\_road\_events}}\) is the number of times the vehicle went off-road, and \(N_{\text{total\_steps}}\) is the total number of episodes. Where \(\text{distance}_i\) is the distance driven during the \(i\)-th safe driving episode, and \(N_{\text{safe\_episodes}}\) is the number of episodes without incidents (such as collisions or off-road events). Where \(N_{\text{completed\_steps}}\) is the number of successfully completed steps, and \(N_{\text{total\_steps}}\) is the total number of steps in the episode. Where \(AR\) is the average reward $f$ collected during the episode.

\begin{table*}[h]
\caption{Hyperparameter settings for the training and evaluation of each baseline}
\label{table1}
\vskip 0.15in
\centering
\resizebox{0.7\textwidth}{!}{%
\begin{tabular}{lcccc}
\toprule
Method & batch size & model size & eval episodes & action repeat \\
\midrule
DDPG & 256 & 32 & 5 & 2 \\
SAC & 256 & 32 & 5 & 2 \\
TD3 & 256 & 32 & 5 & 2 \\
DQN & 256 & 32 & 5 & 2 \\
Latent\_SAC & 256 & 32 & 5 & 2 \\
Dreamer & 256 & 32 & 5 & 2 \\
NFRL & 32 & 32 & 10 & 1 \\
NFRL+SC & 32 & 32 & 10 & 1 \\
BC+Demo & 32 & 32 & 10 & 1 \\
NFRL+SC+Demo & 32 & 32 & 10 & 1 \\
\bottomrule
\end{tabular}%
}
\end{table*}
\begin{table*}[h]
\caption{Hyperparameter settings for the learning rate of each baseline}
\label{table2}
\vskip 0.15in
\centering
\resizebox{0.7\textwidth}{!}{%
\begin{tabular}{lccc}
\toprule
Method & model learning rate & actor learning rate & value learning rate \\
\midrule
DDPG & $1 \times 10^{-4}$ & $3 \times 10^{-4}$ & $3 \times 10^{-4}$ \\
SAC & $1 \times 10^{-4}$ & $3 \times 10^{-4}$ & $3 \times 10^{-4}$ \\
TD3 & $1 \times 10^{-4}$ & $3 \times 10^{-4}$ & $3 \times 10^{-4}$ \\
DQN & $1 \times 10^{-4}$ & $3 \times 10^{-4}$ & $3 \times 10^{-4}$ \\
Latent\_SAC & $1 \times 10^{-4}$ & $3 \times 10^{-4}$ & $3 \times 10^{-4}$ \\
Dreamer & $1 \times 10^{-3}$ & $8 \times 10^{-5}$ & $8 \times 10^{-5}$ \\
NFRL & $1 \times 10^{-3}$ & $8 \times 10^{-5}$ & $8 \times 10^{-5}$ \\
NFRL+SC & $1 \times 10^{-3}$ & $8 \times 10^{-5}$ & $8 \times 10^{-5}$ \\
BC+Demo & $1 \times 10^{-3}$ & $8 \times 10^{-5}$ & $8 \times 10^{-5}$ \\
NFRL+SC+Demo & $1 \times 10^{-3}$ & $8 \times 10^{-5}$ & $8 \times 10^{-5}$ \\
\bottomrule
\end{tabular}%
}

\end{table*}
\subsection{Hyperparameter settings}
\label{sec:appendixA6}
$\mathcal M_{model}$, the KL regularizer is clipped below 3.0 free nats for imagination range $H =15$ using the same trajectories for updating action and value models separately with $\lambda =0.99$ and $\lambda=0.95$, while $k=1.5$. The size of all our trainig and evaluating images is $128\times128\times3$. A random seed $S=5$ is used to collect datasets for the $ego$ vehicle before updating the model every $C=100$ steps during training process. We present the hyperparameter settings for the methods
involved in our experiments as shown in Table8 and Table 9.

\subsection{The world model reconstructs the input images from the original sensors}
\label{sec:appendixA7}
We explores the differences between input images from original sensors and the corresponding reconstructed input images from a world model for 8 types of input. As shown in Figure 7, multiple comparisons are made between the reconstructed input types generated by the world model and their corresponding original sensor inputs. Among them, multi-fusion2, lidar\_noground, lidar+camera and lidar reconstructions are very clear and highly consistent, indicating that $q(o_t|s_t)$ has a precise decoding capability without causing loss of $s_t$. However, birdeye, semantic, (lidar\_noground and multi-fusion3), and (lidar\_noground and camera\_gray) of reconstructions are not as clear as their sensor input. This suggests that world model have difficulty understanding large amounts of irrelevant information related to driving tasks resulting in unclear reconstruction outputs. 

\begin{figure*}[h]
\centering{\includegraphics[width=12cm]{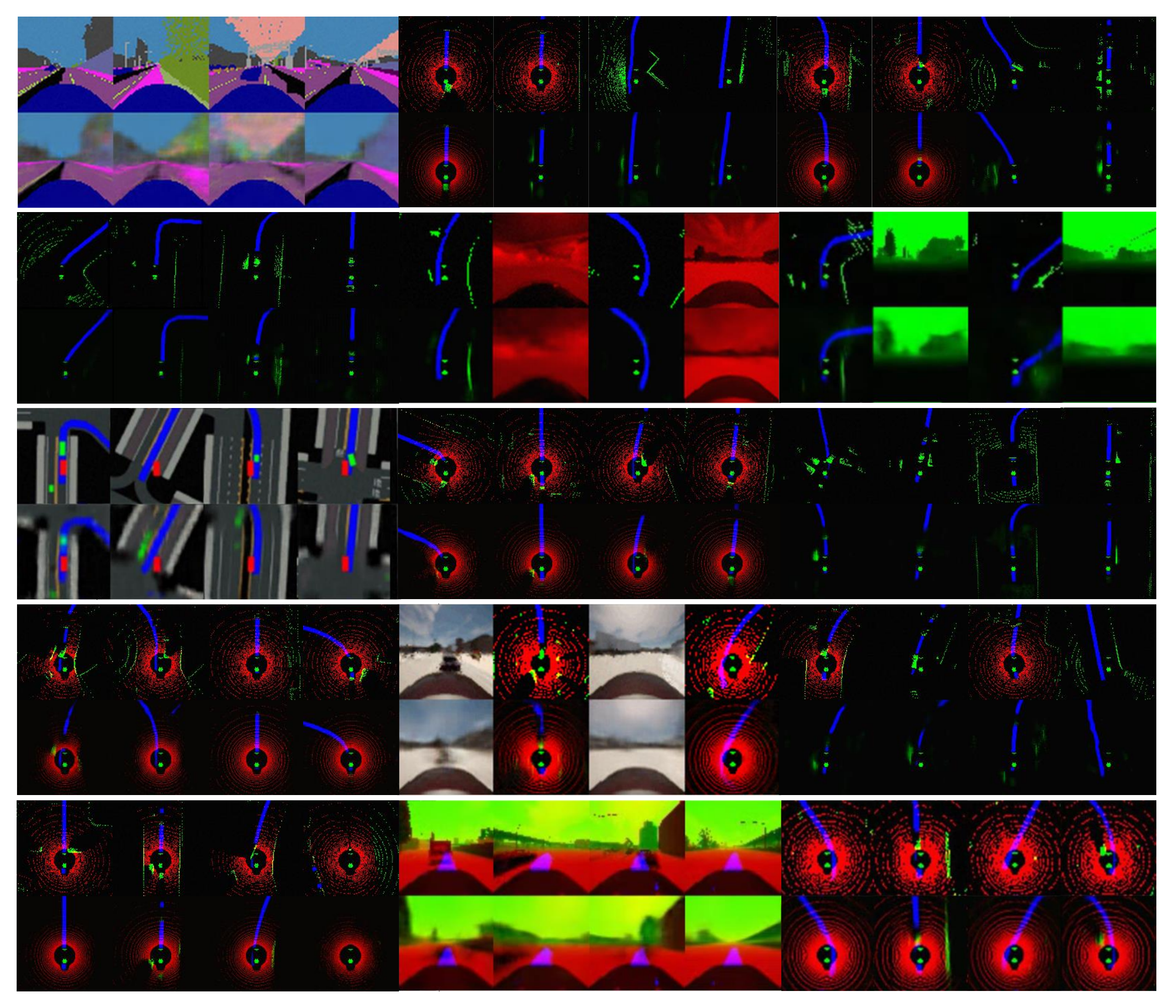}}
\caption{Randomly sampled frames to reconstruct the input images from the original sensors of EGADS on 8 types of input. For each type of image, first row: original sensor inputs. Second row: reconstructed images. } \label{image}
\end{figure*}

\subsection{More results regarding predictions of future driving trajectories}
\label{sec:appendixA8}
The accurate prediction of future driving trajectories is a precondition for making optimal decision making. Random samples of driving trajectories for the first 15 time steps were collected from the sensor. Subsequently, the model predicted the driving trajectories for the next 15 time steps, and the ground truths for these trajectories were also provided We provide additional results regarding predictions of future driving trajectories as shown in
Figure 8 \dots Figure 11.

\begin{figure*}
  \centering
  \begin{subfigure}[b]{\textwidth}
    \includegraphics[width=\linewidth]{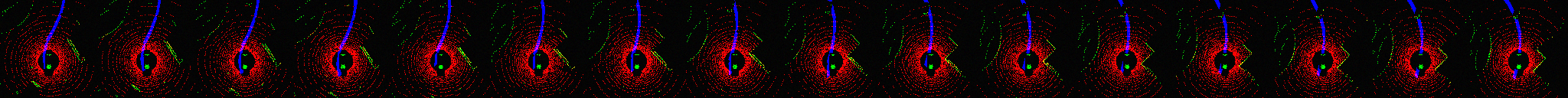}
    \caption{Randomly sample ground truth of inputs Lidar $o_1,o_2,\cdots,o_{15}$}
    \label{fig:sub1}
  \end{subfigure}
  \hfill
  \begin{subfigure}[b]{\textwidth}
    \includegraphics[width=\linewidth]{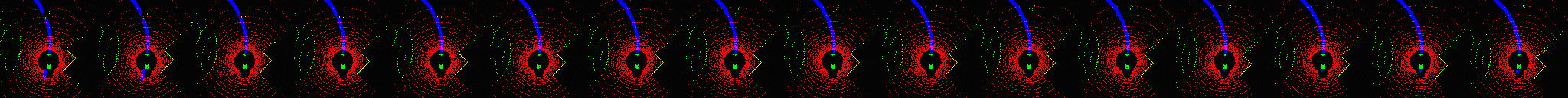}
    \caption{Randomly sample ground truth of inputs Lidar $o_{16},o_{17},\cdots,o_{30}$}
    \label{fig:sub2}
  \end{subfigure}
  \vskip\baselineskip  
  \begin{subfigure}[b]{\textwidth}
    \includegraphics[width=\linewidth]{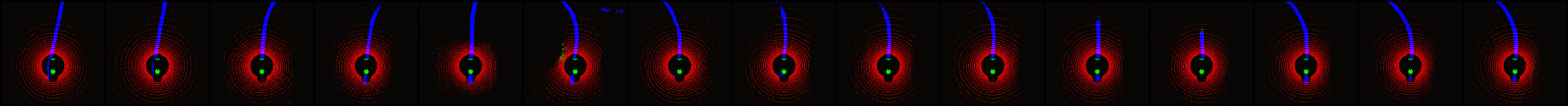}
    \caption{Our model can imagine driving behaviors $\hat{o}_{16},\hat{o}_{17},\cdots,\hat{o}_{30}$}
    \label{fig:sub3}
  \end{subfigure}
  \caption{We randomly sampled input images, and then EGADS was used to make predictions}
  \label{fig:main1}
\end{figure*}

\begin{figure*}
  \centering
  \begin{subfigure}[b]{\textwidth}
    \includegraphics[width=\linewidth]{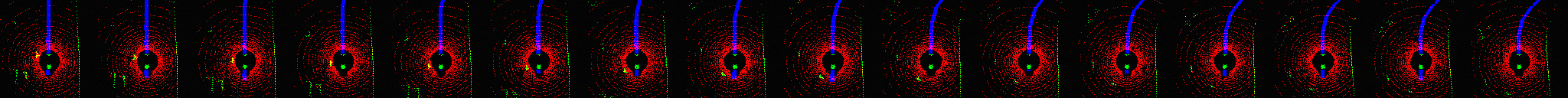}
    \caption{Randomly sample ground truth of inputs Lidar $o_1,o_2,\cdots,o_{15}$}
    \label{fig:sub4}
  \end{subfigure}
  \hfill
  \begin{subfigure}[b]{\textwidth}
    \includegraphics[width=\linewidth]{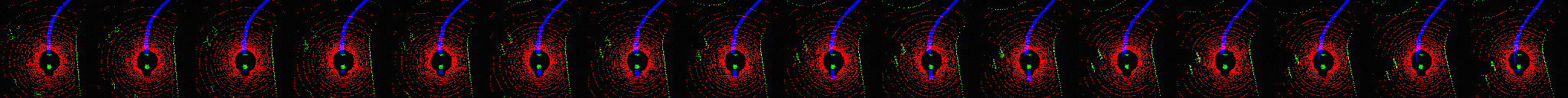}
    \caption{Randomly sample ground truth of inputs Lidar $o_{16},o_{17},\cdots,o_{30}$}
    \label{fig:sub5}
  \end{subfigure}
  \vskip\baselineskip  
  \begin{subfigure}[b]{\textwidth}
    \includegraphics[width=\linewidth]{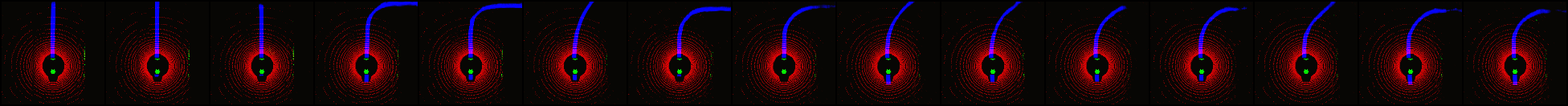}
    \caption{Our model can imagine driving behaviors $\hat{o}_{16},\hat{o}_{17},\cdots,\hat{o}_{30}$}
    \label{fig:sub6}
  \end{subfigure}
  \caption{We randomly sampled input images, and then EGADS was used to make predictions}
  \label{fig:main2}
\end{figure*}

\begin{figure*}
  \centering
  \begin{subfigure}[b]{\textwidth}
    \includegraphics[width=\linewidth]{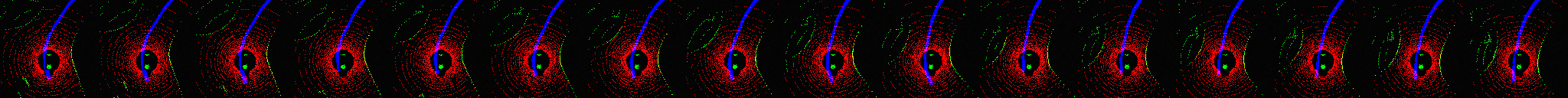}
    \caption{Randomly sample ground truth of inputs Lidar $o_1,o_2,\cdots,o_{15}$}
    \label{fig:sub7}
  \end{subfigure}
  \hfill
  \begin{subfigure}[b]{\textwidth}
    \includegraphics[width=\linewidth]{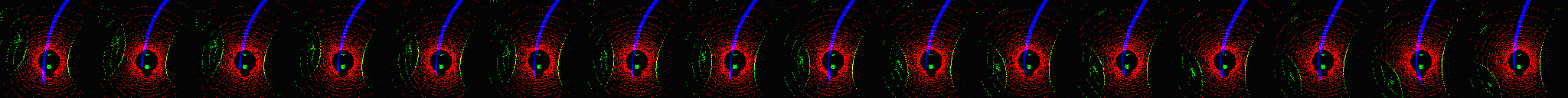}
    \caption{Randomly sample ground truth of inputs Lidar $o_{16},o_{17},\cdots,o_{30}$}
    \label{fig:sub8}
  \end{subfigure}
  \vskip\baselineskip  
  \begin{subfigure}[b]{\textwidth}
    \includegraphics[width=\linewidth]{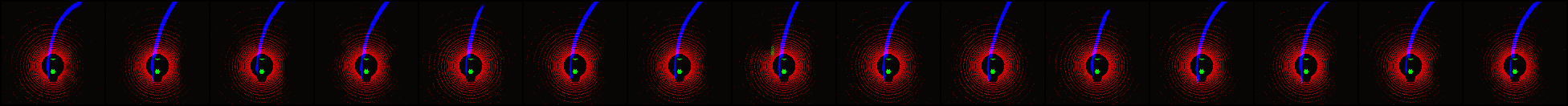}
    \caption{Our model can imagine driving behaviors $\hat{o}_{16},\hat{o}_{17},\cdots,\hat{o}_{30}$}
    \label{fig:sub9}
  \end{subfigure}
  \caption{We randomly sampled input images, and then EGADS was used to make predictions}
  \label{fig:main3}
\end{figure*}

\begin{figure*}
  \centering
  \begin{subfigure}[b]{\textwidth}
    \includegraphics[width=\linewidth]{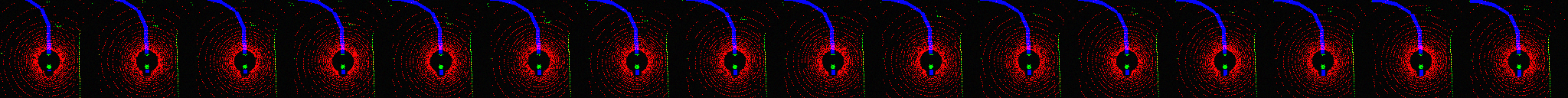}
    \caption{Randomly sample ground truth of inputs Lidar $o_1,o_2,\cdots,o_{15}$}
    \label{fig:sub10}
  \end{subfigure}
  \hfill
  \begin{subfigure}[b]{\textwidth}
    \includegraphics[width=\linewidth]{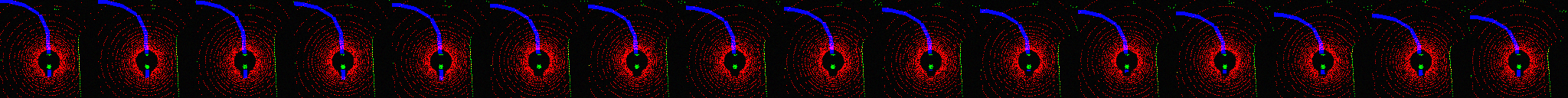}
    \caption{Randomly sample ground truth of inputs Lidar $o_{16},o_{17},\cdots,o_{30}$}
    \label{fig:sub11}
  \end{subfigure}
  \vskip\baselineskip  
  \begin{subfigure}[b]{\textwidth}
    \includegraphics[width=\linewidth]{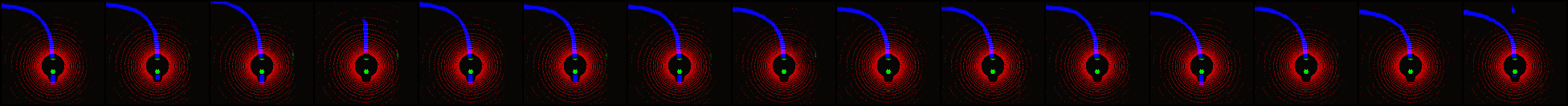}
    \caption{Our model can imagine driving behaviors $\hat{o}_{16},\hat{o}_{17},\cdots,\hat{o}_{30}$}
    \label{fig:sub12}
  \end{subfigure}
  \caption{We randomly sampled input images, and then EGADS was used to make predictions}
  \label{fig:main4}
\end{figure*}


\end{document}